\documentclass[letterpaper, 10 pt, twoside, journal]{IEEEtran}
\usepackage{pdfpages}
\usepackage[noadjust]{cite}
\usepackage{float}
\usepackage{graphicx}
\usepackage{makecell}
\usepackage{amsmath}
\usepackage[ruled,vlined]{algorithm2e}
\usepackage{algpseudocode}
\usepackage{setspace}
\usepackage{booktabs}
\usepackage{multirow}
\usepackage{tabularx}
\usepackage{placeins}
\usepackage{etoolbox}
\usepackage{url}
\usepackage{hyperref}

\AtBeginEnvironment{tabular}{\small}

\newcounter{taggedEquations}
\makeatletter
\def\eqref#1{{\textup{\measuring@true\tagform@{\ref{#1}}}}}%
\def\maketag@@@#1{\hbox{%
\ifmeasuring@\else
  \stepcounter{taggedEquations}%
\fi
\m@th\normalfont#1}}
\makeatother

\usepackage{tikz}
\usepackage{textcomp}
\usepackage{lipsum}

\newcommand\copyrighttext{
  \footnotesize \textcopyright © 2022 IEEE.  Personal use of this material is permitted.  Permission from IEEE must be obtained for all other uses, in any current or future media, including reprinting/republishing this material for advertising or promotional purposes, creating new collective works, for resale or redistribution to servers or lists, or reuse of any copyrighted component of this work in other works.
  DOI: \href{https://ieeexplore.ieee.org/document/9935063}{10.1109/LRA.2022.3219018}}
\newcommand\copyrightnotice{
\begin{tikzpicture}[remember picture,overlay]
\node[anchor=south,yshift=0pt] at (current page.south) {\fbox{\parbox{\dimexpr\textwidth-\fboxsep-\fboxrule\relax}{\copyrighttext}}};
\end{tikzpicture}%
}

\begin{document}

\title{\LARGE \bf
Point Cloud Change Detection With Stereo V-SLAM: \protect\\ Dataset, Metrics and Baseline
}

\author{Zihan Lin$^{1}$, Jincheng Yu$^{1}$, Lipu Zhou$^{2}$, Xudong Zhang$^{1}$, Jian Wang$^{1}$, Yu Wang$^{1}$
\thanks{Manuscript received: June 28, 2022; Revised September, 26, 2022; Accepted October 20, 2022.}
\thanks{This paper was recommended for publication by Editor Sven Behnke. Name upon evaluation of the Associate Editor and Reviewers' comments.}
\thanks{
This work was supported by  National Natural Science Foundation of China (U20A20334, U19B2019 and M-0248), Tsinghua-Meituan Joint Institute for Digital Life, Tsinghua EE Independent Research Project, Beijing National Research Center for Information Science and Technology (BNRist), and Beijing Innovation Center for Future Chips.} 
\thanks{$^{1}$Department of Electronic Engineering,
        Tsinghua University, Beijing, 100084, China. jian-wang@tsinghua.edu.cn}
\thanks{$^{2}$Meituan, 7 Rongda Road, Chaoyang District, Beijing, 100012, China.}
\thanks{Corresponding author: Jian Wang}
\thanks{Our code is in the repository: \protect\url{https://github.com/lnexenl/PPCA-VINS.git}, and dataset is also released on this page.} 
}

\markboth{IEEE Robotics and Automation Letters. Preprint Version. Accepted OCTOBER, 2022}%
{Lin \MakeLowercase{\textit{et al.}}: Point Cloud Change Detection With Stereo V-SLAM: Dataset, Metrics and Baseline}

\maketitle

\copyrightnotice
\begin{abstract}
Localization and navigation are basic robotic tasks requiring an accurate and up-to-date map to finish these tasks, with crowdsourced data to detect map changes posing an appealing solution. Collecting and processing crowdsourced data requires low-cost sensors and algorithms, but existing methods rely on expensive sensors or computationally expensive algorithms. Additionally, there is no existing dataset to evaluate point cloud change detection. Thus, this paper proposes a novel framework using low-cost sensors like stereo cameras and IMU to detect changes in a point cloud map. Moreover, we create a dataset and the corresponding metrics to evaluate point cloud change detection with the help of the high-fidelity simulator Unreal Engine 4. Experiments show that our visual-based framework can effectively detect the changes in our dataset.

\end{abstract}


\section{INTRODUCTION}
\IEEEPARstart{A}{utonomous} ground and unnamed aerial vehicles have become popular over the last few years. These robotic vehicles are commonly used in various tasks, such as logistics delivery and remote sensing, with localization and navigation being mandatory sub-tasks requiring an accurate and up-to-date map. Thus, detecting the map changes is crucial.

This work {is designed to} exploit crowdsourced data to detect map changes in real-time. Crowdsourced data is collected from several vehicles equipped with sensors like a camera, and the location where vehicles pass also contributes to detecting map changes. When many vehicles collect data, the overall sensor cost and the collected data are enormous. Hence, these vehicles should be equipped with low-cost sensors to reduce hardware costs, and crowdsourced data should be processed using an algorithm with low computational complexity.

Current change detection methods utilize an expensive hardware setup. For example, remote sensing tasks utilize expensive sensors like vehicle-based laser scanner (VLS), airborne laser scanner (ALS), and mobile laser scanner (MLS) to detect point cloud changes \cite{vls_rs,als_rs,mls_rs}. However, the price of a laser scanner ranges from thousands to millions of dollars. Alternative solutions exploit the structure from motion (SfM), and multi-view stereo (MVS) technology \cite{city_scale_pccd}, but these methods impose a huge computational complexity, typically requiring several hours to days of calculations, depending on the point cloud size. For example, Yew and Lee \cite{city_scale_pccd} developed a SfM-based algorithm, where the reconstruction and registration steps need three days for a medium-size urban environment.

Considering the available datasets, EuRoC \cite{euroc} and KITTI \cite{kitti}, and their corresponding metrics focus on evaluating the performance of localization and mapping. Moreover, the SHREC 2021 point cloud change detection dataset \cite{shrec} focuses on change detection of point clouds collected by LiDAR. Hence, to the best of our knowledge, datasets and metrics to evaluate visual-based point cloud change detection do not exist, and thus evaluating the performance of change detection is a challenging task.

Hence, this paper proposes a novel visual-based framework to achieve low-cost change detection in point clouds and a dataset with the appropriate metrics to evaluate the performance of point cloud change detection. The main contributions of this paper are as follows:

\begin{figure}[h]
        \centering
        \includegraphics[width=0.5\textwidth]{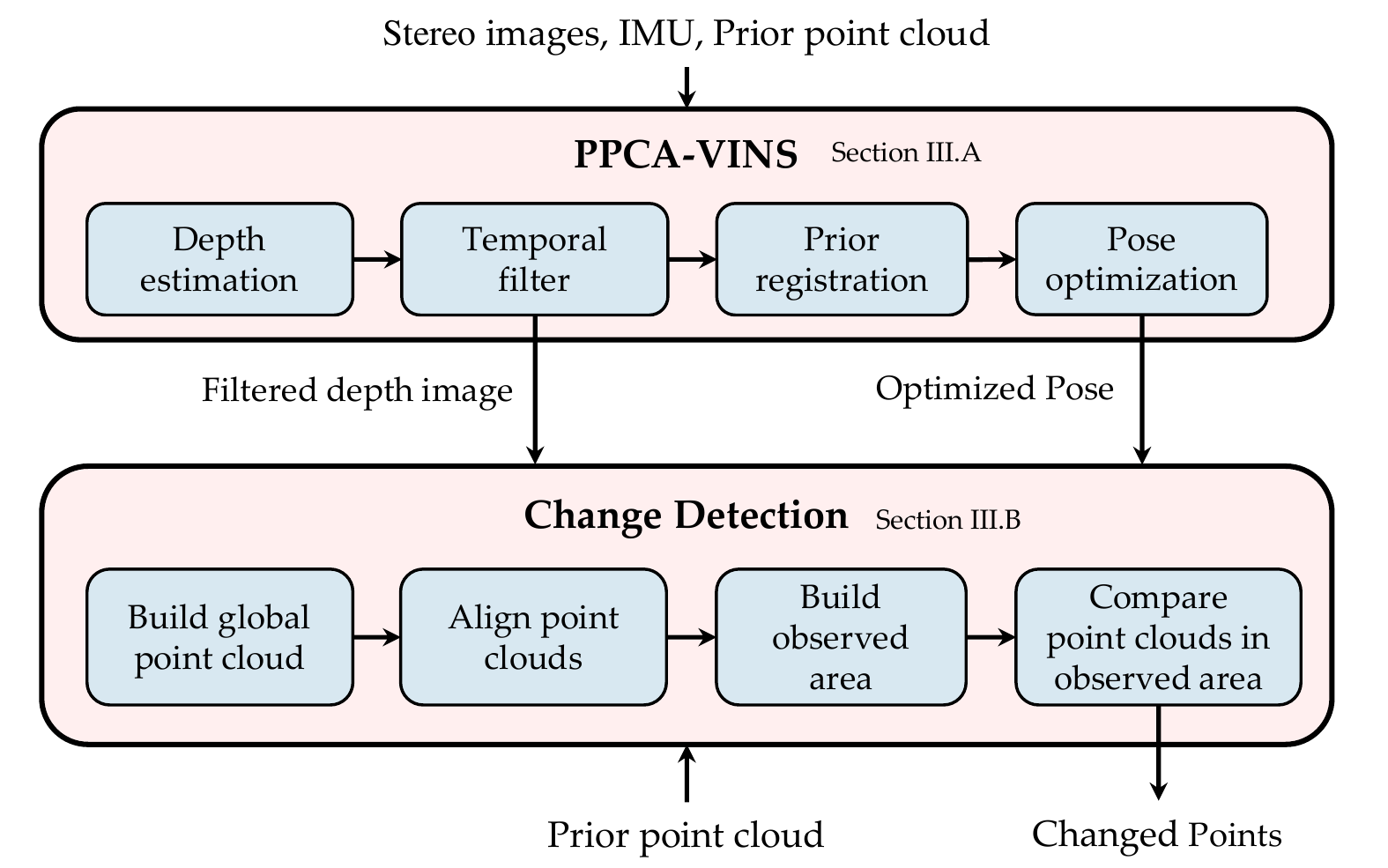}
        \caption{Proposed change detection framework comprising the PPCA-VINS and Change Detection modules. PPCA-VINS provides 6D poses and depth images for change detection, which are then input to the change detection pipeline to build a global point cloud. Then the global point cloud and the prior point clouds of an observed area are compared, with the change detection pipeline providing the point cloud vertices that have changed.}
        \label{pipeline}
\end{figure}

\begin{itemize}
        \item Proposing a prior point cloud-assisted SLAM framework named PPCA-VINS (prior point cloud-assisted VINS), based on VINS-Fusion \cite{vins_fusion} and assuming having an accurate initial pose. Compared to standard VINS-Fusion, our frame fuses prior information from prior point cloud and affords an appealing performance over the standard VINS-Fusion on our dataset. (A prior point cloud is a point cloud of the scene that vehicle is in, but the prior point cloud is out-dated and needs change detection.)
        \item Designing a new \textit{dataset and the corresponding metrics} for visual-based point cloud change detection tasks. The dataset fills the vacancy in the point cloud change detection task and provides raw data from multiple sensor types onboard a car or an unnamed aerial vehicle.
        \item Developing a visual-based change detection architecture that utilizes depth images and poses provided by PPCA-VINS to build a global point cloud. We detect changes by comparing the global point cloud to the prior point cloud. Our change detection pipeline is evaluated on our change detection dataset, verifying its capability for point cloud change detection.
\end{itemize}

The proposed framework is illustrated in Fig.\ref{pipeline} and comprises two parts. The first part is PPCA-VINS (Section \ref{cross-modal}), which estimates depth from stereo images and then builds filtered depth images. Then we convert the filtered depth image to a point cloud and register it to the previous point cloud. Finally, we optimize the vehicle's pose by fusing the pose obtained from VINS-Fusion and the prior registration. 

The second part of our architecture is the change detection module (Section \ref{change-detection}). In this module, we first build a global point cloud using the filtered depth images and optimized poses from PPCA-VINS, and then we align the global point cloud to the prior point clouds. Then we build the point cloud of the observed area and compare the global point cloud to the prior point clouds using point cloud registration to obtain the changed vertices.

The remainder of this paper is as follows: Section II presents the related work, while Section III includes the proposed method. Section \ref{dataset} builds the suggested dataset using the high-fidelity simulator Unreal Engine 4 \cite{unrealengine} due to ease of changing scene and data collection. Moreover, this section suggests some evaluation metrics. Section \ref{experiment} conducts several experiments to demonstrate the framework's and dataset's effectiveness on map change detection tasks using low-cost sensors and algorithms. Finally, section VI concludes this work.

\section{RELATED WORK}
\subsection{Visual Localization with Prior Map}
Visual localization has been researched for several years.  For example, Zuo \textit{et al.} \cite{msckf_prior} proposed a tightly-coupled {Multi-State Constraint Kalman Filter (MSCKF) \cite{msckf} scheme using stereo vision, where MSCKF is a SLAM framework employing Kalman filter and sliding window.} In this work, the local point clouds were generated from a stereo camera setup and utilized {Normal Distribution Transform (NDT)} \cite{ndt} to align the local point clouds and the prior LiDAR map, where the location in the prior map was considered as a state in the MSCKF's extended Kalman filter. Lynen \textit{et al.} \cite{get_out_of_my_lab} developed a pipeline based on SfM. This method was also MSCKF-based and tightly coupled. The difference between \cite{get_out_of_my_lab} and  \cite{msckf_prior} is the prior localization algorithm, as \cite{get_out_of_my_lab}  employed SfM to generate the prior point cloud map and stored the visual landmark descriptors as a database in the meantime. The descriptors obtained from the images were calculated and matched against the database landmarks, and then the pose relationship between the 2D feature point and 3D landmarks was filtered using an extended Kalman filter to obtain the final estimated pose. 
{Previous works localizes using image sequences from a camera as temporal information, but the following work provides localization result with only single image.  
Li and Lee} \cite{deepi2p} suggested a deep learning method named DeepI2P to register images to point clouds. In this work, the authors converted the cross-modal registration problem into a classification problem, where a deep neural network is trained to classify whether a point in a point cloud is within the image frustum. In that case, a post-optimization method is applied to estimate the images' poses.

\subsection{Change Detection}

{Yew and Lee} \cite{city_scale_pccd} combined SfM and neural networks, where the SfM built a global point cloud and the neural network performed non-rigid registration between the point clouds. The author claimed that their algorithm takes several days to calculate, which highlights that SfM-based methods are extremely time-consuming. In addition, this method did not distinguish the unknown and empty areas. Xu \textit{et al.} \cite{octree_cd} utilized ALS and octrees. Specifically, this method collected point clouds utilizing ALS  and the octree stored and indexed the irregularly-distributed points. Then, the points were clustered and classified into different kinds, followed by a registration process to reveal the changed areas. Although this work did not distinguish between unknown and empty areas, it is reasonable to assume that collecting LiDAR data from airborne sensors do not produce many unknown areas. 

{Besides from change detection with point cloud, there are some other image-based techniques to detect changes from images. Sakurada \textit{et al.} \cite{6618869} propose a probabilistic model to detect changes. Different from \cite{city_scale_pccd}, they build a probabilistic model to represent the structure of an outdoor scene and evaluate the probability of changes. Sakurada and Okatani \cite{jst2015change} develop a neural network based method to develop changes between images taken at same place but at different time. They exploit neural network to output image features and segmentation results, and they detect changes in images using these outputs. Taneja \textit{et al.} \cite{taneja2011image} builds a 3D model with multi-view stereo first. Then they collect images from low resolution cameras and detect changes against the 3D model by detecting inconsistencies between images and the model. Taneja \textit{et al.} \cite{taneja2013city} then develops a methods detecting changes in city's cadastral model using images. Based on previous work \cite{taneja2011image}, they propose a new method to detect changes in large scale with sparse imagery.} \\

Our work employs VINS-Fusion, a unified framework for pose estimation, to provide the VIO pose for our pipeline. 
{VINS-Fusion is a general optimization-based SLAM framework which can fuse multiple sensors (stereo cameras, IMU, Magnetometer etc.). First, VINS-Fusion calculates 6D VIO poses with inputs from cameras and IMU. Then a global pose optimization is made to fuse 6D VIO poses and inputs from other sensors, and finally VINS-Fusion outputs global 6D poses 
 after optimization. Besides, VINS-Fusion uses keyframe technique, a keyframe is a frame containing timestamp, images, IMU readings and a 6D pose, and there should be enough parallax between two keyframes' images.}
 
 In addition, VINS-Fusion use a sliding window to reduce processing time. Thus a tightly-coupled VIO requires a quick registration between the local and the prior point clouds—otherwise, the keyframe is discarded from the sliding window. Opposing current works, the proposed method relies exclusively on cheap sensors like stereo camera, GPS and IMU. Moreover, our pipeline is computationally efficient; thus, our entire framework is low-cost.

\section{FRAMEWORK}
\subsection{PPCA-VINS}\label{cross-modal}
\subsubsection{Depth Estimation}
VINS-Fusion can track feature points and generate a sparse point cloud from these points. However, using a sparse point cloud does not provide an accurate localization result through registration. Thus, additional work producing a semi-dense point cloud is expected.

One standard method of cross-modal data processing is converting different modalities into one. In our pipeline, we convert stereo images into point clouds, as stereo depth estimation has been a popular research topic for several years. Semi-global matching (SGM) \cite{sgm} is a classic method for disparity estimation, based on minimizing the mutual information (MI) between two images. Recently, many neural networks (NNs) based on stereo depth estimation methods have emerged, such as PSMNet \cite{PSMNet} and AnyNet \cite{anynet}. However, these methods have a significant time complexity, and the depths produced are not better than the traditional SGM method. Thus, we leverage SGM to produce depth estimation for stereo images.

\subsubsection{Temporal Filter}
The depth images produced by SGM incorporate significant noise. Thus, we adopt the temporal filter proposed in  \cite{kinect_temporal_filter} and a similar method named \textit{depth correspondence matching} \cite{msckf_prior}. We follow these methods and implement a simplified version of the temporal filter. 

Let the current keyframe be $i$. We re-project the estimated depth of every pixel $(u,v)$ in keyframe $j \ (i-2\leq j \ \leq i+2, j\neq i)$ to the current keyframe $i$ and obtain the projected depth $d_j^i(u, v)$ and pixel coordinate $(u', v')$. If the distance between the projected depth $d_j^i(u, v)$ and the current frame depth $d_i(u', v')$ is smaller than a threshold $\delta_d$, we call this a successful projection. When a pixel is successfully projected over $\alpha$ times, we average the projected depth and the current frame depth of this pixel to obtain a filtered depth, which is added to the filtered depth image.

\subsubsection{Prior Registration}
In this subsection, we register local point clouds to prior point cloud to get prior localization results. We increase the point cloud's density by aggregating the point cloud of every five keyframes to a local cloud, producing a denser point cloud.

For our pipeline, we need to register two point clouds, each including tens of thousands of points. Thus, we need a fast and accurate registration algorithm. {Normal distribution transform (NDT) \cite{ndt} is a registration algoritghm based on normal distribution. It divides points into small cells, and uses normal distribution to represent points' distribution in cells, thus accelerating the computation.} But NDT is unsuitable for our pipeline, as it often produces results suffering from a considerable drift. Alternatively, the iterative closest point (ICP) \cite{icp} is an accurate registration algorithm but it requires much time even to align two small point clouds. Besides, the Generalized ICP (GICP) \cite{gicp} is an improved ICP-based algorithm that balances speed and accuracy. Koide \textit{et al.} \cite{fgicp} boosted GICP by implementing the algorithm on multiple threads.  Indeed, we conducted several tests comparing a 4-threaded GICP and other single threaded algorithms involving a point cloud and its variant translated by 0.2m. { And we found that 4-threaded GICP can run fast and achieve a good result. }

Registration is not always accurate, and therefore we evaluate the registration results using the fitness metric, which is the mean distance from each point in the local cloud to its closest point in the prior cloud. We evaluate every registration based on the following:
\begin{enumerate}
        \item Registration should be converged.
        \item Fitness of registration should be less than a threshold.
        \item Registration's translation length should be less than a threshold.
\end{enumerate}

Note that GICP is formulated as a Maximum-Likelihood Estimation (MLE) problem, so the Hessian matrix in the GICP registration is the negative of the Fisher information matrix. The registration's covariance matrix is the inverse of the Fisher information matrix, written as follows:
\begin{align}\label{cov}
       \texttt{cov}(\theta) = \mathcal{I}^{-1} (\theta)= (-\mathbf{H}_\theta)^{-1}
\end{align}

The covariance matrix is used in the following global pose estimation.

\subsubsection{Pose Optimization} 
 Based on VINS-Fusion, we consider our registration results as outputs from a pseudo sensor. Then we fuse registration results and 6D VIO poses to get global 6D poses.

The prior localization states considering the prior position and orientation of the $k$-th keyframe are:
\begin{align}
        \mathbf p_k^p ,\mathbf q_k^p 
\end{align}
In VINS-Fusion, the global pose graph optimization problem is modelled as a maximum likelihood estimation (MLE) problem. The variables to be optimized are global poses:
\begin{align}
        \mathcal{X} &=[\mathbf{x}_0, \mathbf{x}_1, \cdots \mathbf{x}_n ]\\
        \mathbf x_k &= [\mathbf p_k, \mathbf q_k], k \in [0, n]
\end{align}
Assuming that the measurement uncertainties are Gaussian distributions, the problem is formulated as :
\begin{align}
        \mathcal{X}^* &= \arg \max\limits_\mathcal{X} \prod_{t=0}^n \prod_{k\in \mathbf S} p(\mathbf{z}_t^k|\mathcal{X}) \\
        &= \arg \min\limits_\mathcal{X} \sum_{t=0}^n \sum_{k\in \mathbf S}\lVert \mathbf z_t^k - h_t^k(\mathcal X)\rVert^2_{\Omega_t^k }
\end{align}

Since we use the same VIO factor as in the VINS-Fusion's original paper, we do not explain it here to enhance readability. Considering the prior localization factor, we divide it into two parts. One is the global position factor:
\begin{align}
        \mathbf z_{t,p}^{p} - h_{t,p}^{p}(\mathcal{X}) = \mathbf z_{t,p}^p - h_{t,p}^p(\mathbf x_t) = \mathbf p_{t}^p- \mathbf p_t
\end{align}

and the other one is the relative rotation factor from the initial frame to current frame t:
{
\begin{align}
        \mathbf z_{t,q}^{p} - h_{t,q}^{p}(\mathcal{X}) = \mathbf z_{t,q}^{p} - h_{t,q}^{p}(\mathbf x_1, \mathbf x_t) =({\mathbf q_1^p})^{-1} \mathbf q_t^p \ominus ({\mathbf q_1})^{-1} \mathbf q_t
\end{align}

Here $\mathbf z$ denotes observations from sensors, and $h$ denotes observation function transforming poses to observed values.}

Considering the prior localization factor, we build a pose graph and optimize it to obtain the global pose. The covariance of the prior localization is given in eq.\ref{cov}, which is used to calculate the confidence of the prior localization.

\subsection{Change Detection}\label{change-detection}
This subsection builds a simple change detection algorithm that comprises the following steps:
\begin{enumerate}
        \item build the global point cloud $\mathcal{P}_g$
        \item align the global point cloud to the prior point cloud $\mathcal{P}_p$ using {GICP}
        \item build the observed area $\mathcal{A}_{obs}$
        \item build the observed prior point cloud ($\mathcal{P}_p^{obs}$) and compare $\mathcal{P}_p^{obs}$ and $\mathcal{P}_g$
\end{enumerate}

Initially, we use the keyframes' 6D pose and filtered depth image to build a global point cloud. Then we register the global point cloud to the prior point cloud to align the two clouds. 

\begin{figure}[h]

        \centering
        \includegraphics[width=0.45\textwidth]{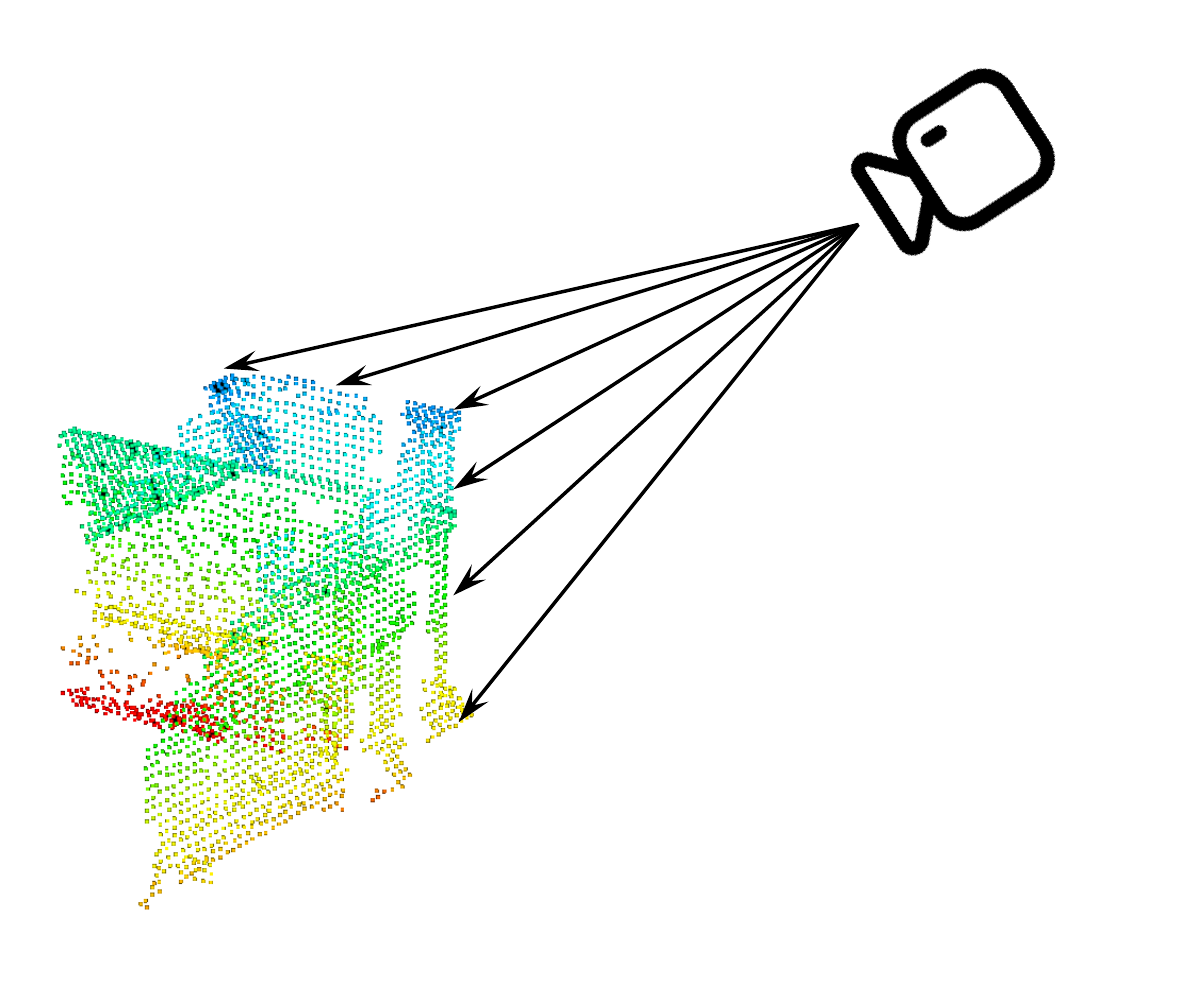}
        \caption{Illustration of how $A_{obs}$ is built. For every keyframe, we have a corresponding point cloud and 6D camera pose. We cast rays from camera to point cloud, and the voxels which rays passed are considered a part of $A_{obs}$}
        \label{ray-casting}
\end{figure}

In step 3, we must know what part of the point cloud has been observed to determine the part of the point cloud that has changed. So we leverage the octomap \cite{octomap} of resolution $\rho_o$ to build an observed area $\mathcal{A}_{obs}$. For the pixels in the filtered depth image with a depth smaller than $th_d$, we cast a ray from the camera to the point. {(Fig.\ref{ray-casting})} For those pixels without a depth, we also cast a ray from the camera with a certain depth $th_f$ to add some free space to octomap. The optimal value of $th_f$ is experimentally defined.

In step 4, we build the observed prior point cloud $\mathcal{P}_p^{obs}$. Given the depth estimation inaccuracy, we use distance to $\mathcal{P}_g$ to judge whether a point of $\mathcal{P}_p$ is observed. So we consider a point of $\mathcal{P}_p$ is observed when it is in $\mathcal{A}_{obs}$ or is within a distance $th_{ch}$ of $\mathcal{P}_g$, as shown in Algorithm \ref{alg:obs}. 

Finally, we compare the point clouds by calculating the point-to-cloud distance between $\mathcal{P}_g$ and $\mathcal{P}_p^{obs}$. If a point in $\mathcal{P}_g$ is not within a distance $th_{ch}$ of $\mathcal{P}_p^{obs}$, we classify this point as a new point. If a point in $\mathcal{P}_p^{obs}$ is not within a distance $th_{ch}$ of $\mathcal{P}_g$, we classify this point as a removed point. Finally, a set of new points $\mathcal{P}_{new}$ and a set of removed points $\mathcal{P}_{rm}$ are built, as shown in Algorithm \ref{alg:classify}.

\begin{algorithm}
\footnotesize
        \caption{Building the observed prior point cloud }\label{alg:obs}
        \KwData{$\mathcal{P}_p, \mathcal{P}_g, \mathcal{A}_{obs}, th_{ch}$}
        \KwResult{Observed prior point cloud $\mathcal{P}_p^{obs}$}
        \For{$p \in \mathcal{P}_p$}{
          \If{$p \in \mathcal{A}_{obs} \ \textbf{or} \ calcNearestDist(pt, \mathcal{P}_g) \le th_{ch}$}{
                $\mathcal{P}_p^{obs}.append(p)$
          }
        }
\end{algorithm}

\begin{algorithm}
\footnotesize
\caption{Classification of points}\label{alg:classify}
\KwData{$\mathcal{P}_g, \mathcal{P}_p^{obs}$}
\KwResult{$\mathcal{P}_{new}, \mathcal{P}_{rm}$}
\For{$p \in \mathcal{P}_g$}{
  \If{$calcNearestDist(p, \mathcal{P}_p^{obs}) \geq th_{ch}$}{
        $\mathcal{P}_{new}.append(p)$
  }
}
\For{$p \in \mathcal{P}_p^{obs}$}{
  \If{$calcNearestDist(p, \mathcal{P}_g) \geq th_{ch}$}{
        $\mathcal{P}_{rm}.append(p)$
  }
}
\end{algorithm}


\section{DATASET AND METRICS}\label{dataset}
\subsection{Dataset}

\begin{figure}[!h]
        \centering
        \includegraphics[width=0.5\textwidth]{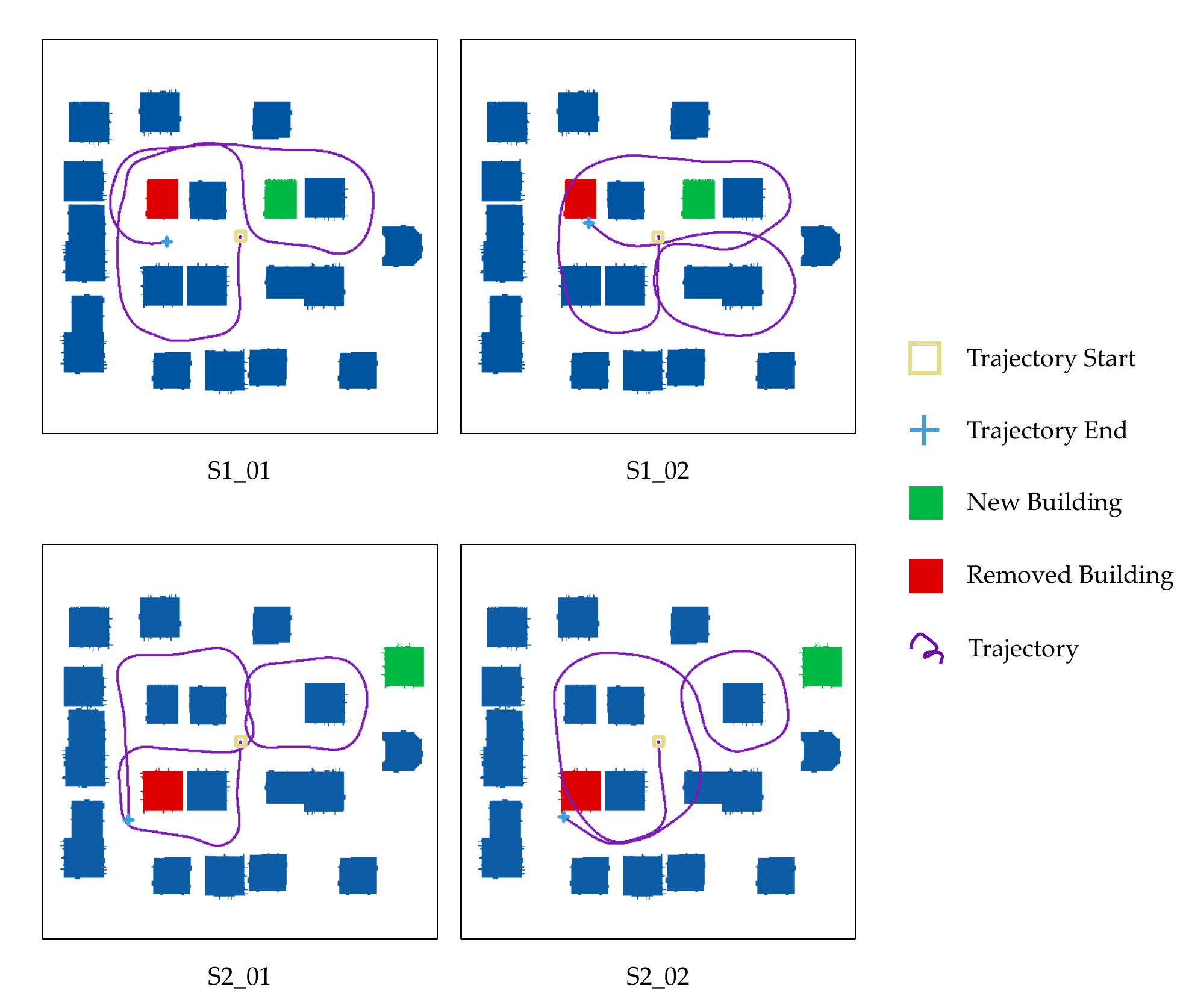}
        \caption{Top view of scenes and trajectories.}
        \label{fig:traj}
\end{figure}

To evaluate our change detection pipeline, we build a dataset using the simulated scene in Unreal Engine 4 with the AirSim \cite{airsim} plugin. The simulated dataset affords quickly changing scenes and collecting data in a simulated world. {We have built three scenes in good light condition, namely the original scene and two changed scenes (S1, S2, and in each one we remove a building and add a new building (Fig.\ref{fig:traj})).} Each scene has a size of 250$\times$250$\times$40 meters (length$\times$width$\times$height), {and for each scene, we also build its ground truth point cloud using LiDAR. The groundtruth point cloud of original scene is used as prior point cloud.}

In each scene, we record 2 MAV's trajectory (Fig.\ref{fig:traj}) {at different height (4m and 15m)} with 30 Hz 800$\times$450 stereo pairs, 200 Hz IMU and ground truth 6D pose data. The stereo pairs and IMU are noiseless. {All these trajectories have an approximately 170s duration and a length about 700 meters. And we name these trajectories using scene name and height. (e.g. "S1\_01", "S2\_02", "S1" means this trajectory is collected in S1 scene, and "01" means this trajectory is recorded at a height about 4m, "02" means this trajectory is recorded at a height about 15m)}

{Except for these 4 trajectories, we also collected 2 trajectories with more difficulties: S1\_mirror with mirrors in the scene and S1\_dark with bad lighting condition. For S1\_mirror, we place one mirror on the outer-wall of one building and the other mirror on the ground of removed building in S1 scene. For S1\_dark, we decrease sun light and sun height, so it's much darker than original S1 scene. These two trajectories are with the same trajectory as S1\_01.}

\begin{algorithm}[!h]
    \footnotesize
        \setstretch{1.4}
        \caption{Build clouds from the true positive points}\label{algo:tp}
        \KwData{$\mathcal{P}_{p,rm}^{obs},\mathcal{P}_{p,new}^{obs},\mathcal{P}_{rm},\mathcal{P}_{new}$}
        \KwResult{$\mathcal{P}_{p,rm}^{obs,TP},\mathcal{P}_{p,new}^{obs,TP},\mathcal{P}_{rm}^{TP},\mathcal{P}_{new}^{TP}$}
        \For{$p \in \mathcal{P}_{p,rm}^{obs}$}{
                \If{$calcNearestDist(p, \mathcal{P}_{rm}) \leq th_{ch}$}{
                        $\mathcal{P}_{p,rm}^{obs,TP}.append(p)$
                }
        }
        \For{$p \in \mathcal{P}_{p,new}^{obs}$}{
                \If{$calcNearestDist(p, \mathcal{P}_{new}) \leq th_{ch}$}{
                        $\mathcal{P}_{p,new}^{obs,TP}.append(p)$
                }
        }
        \For{$p \in \mathcal{P}_{rm}$}{
                \If{$calcNearestDist(p, \mathcal{P}_{p,rm}^{obs}) \leq th_{ch}$}{
                        $\mathcal{P}_{rm}^{TP}.append(p)$
                }
        }
        \For{$p \in \mathcal{P}_{new}$}{
                \If{$calcNearestDist(p, \mathcal{P}_{p,new}^{obs}) \leq th_{ch}$}{
                      $\mathcal{P}_{new}^{TP}.append(p)$
                }
        }
\end{algorithm}

\subsection{Metrics}\label{metrics}

In our pipeline, the changed area is presented as a point cloud. Hence, we consider the change detection problem as a classification problem of points, and therefore, we introduce some standard classification metrics to the change detection problem.

We employ the following metrics for change detection evaluation:
\begin{enumerate}
        \item Recall of class new: $R_{new}$
        \item Precision of class new: $P_{new}$
        \item Recall of class removed: $R_{rm}$
        \item Precision of class removed: $P_{rm}$
\end{enumerate}

Besides, we denote $N(\cdot)$ as the number of points in a point cloud.

To evaluate the algorithms on the same standards, all point clouds are downsampled to a certain resolution $\rho_p$. In our paper, $\rho_p=0.4 m$.

We must know the prior (i.e. groundtruth) new and the removed points to calculate these metrics. Hence, in our change detection algorithm, we use the same algorithm as in Algorithm \ref{alg:classify}. Specifically, we replace $\mathcal{P}_g, \mathcal{P}_p^{obs}$ with the prior point cloud of the changed and the original scene, and the algorithm outputs prior new points $\mathcal{P}_{p,new}$ and the prior removed points $\mathcal{P}_{p,rm}$ of the changed scene. 

Similarly, for the change detection algorithms, we must know which points are observed in the prior new points and prior removed points. To obtain these points, we replace $\mathcal{P}_p$ with $\mathcal{P}_{p,new}$ and $\mathcal{P}_{p,rm}$  in Algorithm\ref{alg:obs} and obtain the observed prior new and removed points $\mathcal{P}_{p,new}^{obs}$, $\mathcal{P}_{p,rm}^{obs}$.

Superscript "TP" marks the point cloud with correctly classified points. For example, $\mathcal{P}_{rm}^{TP}$ means  points in $\mathcal{P}_{rm}$ that are within a distance $th_{ch}$ of $\mathcal{P}_{p,rm}^{obs}$, as presented in Algorithm \ref{algo:tp}.

Thus the metrics can be written as:
\begin{align}
R_{new} = \cfrac{N(\mathcal{P}_{p,new}^{obs,TP})}{N(\mathcal{P}_{p,new}^{obs})},\ \ P_{new} = \cfrac{N(\mathcal{P}_{new}^{TP})}{N(\mathcal{P}_{new})} \\
R_{rm}  = \cfrac{N(\mathcal{P}_{p,rm}^{obs,TP})}{N(\mathcal{P}_{p,rm}^{obs})},\ \ P_{rm}  = \cfrac{N(\mathcal{P}_{rm}^{TP})}{N(\mathcal{P}_{rm})} 
\end{align}

Besides, we challenge our dataset against other point cloud change detection datasets like the SHREC 2021 \cite{shrec} and the dataset built by Gélis \textit{et al.} \cite{cdupc}. The corresponding results are reported in Table \ref{dataset_comp}.
\AtBeginEnvironment{tabular}{\scriptsize}
\begin{table}[h]
        \caption{Comparison of Point Cloud Change Detection Datasets}
        {\scriptsize
        \label{dataset_comp}
        \begin{center}
        \begin{tabular}{ccc}
        \toprule
        Dataset & Scene type & Sensor \\
        \midrule
        Ours & simulated urban & Stereo camera, IMU \\
        SHREC 2021 \cite{shrec} & real urban & LiDAR \\
        Gélis \textit{et al.} \cite{cdupc} & simulated urban & LiDAR \\
        \bottomrule\\
        \toprule
        Dataset & Vehicle type & Metrics \\
        \midrule
        Ours & aerial & \makecell[c]{recall, precision \\of points' classes}\\
        \midrule
        SHREC 2021 & ground & \makecell[c]{accuracy, IoU \\of objects' classes}\\
        \midrule
        Gélis \textit{et al.} & aerial & \makecell[c]{ IoU \\of points' classes} \\
        \bottomrule
        \end{tabular}
        \end{center}
        }
\end{table}

\section{EXPERIMENTAL RESULTS}\label{experiment}
Our experiments can be divided into two parts: First, the prior localization test shows our PPCA-VINS's advantage over no prior information-assisted methods. Second, the change detection test demonstrates that our change detection dataset and metrics can effectively evaluate a change detection algorithm and that our visual-based baseline pipeline works successfully on the dataset.In addition, in our experiment we assume that we have an accurate initial guess of the vehicle's pose. All experiments are conducted on a desktop PC with an Intel i9-11900K, an Nvidia RTX 3090, and 32 GB RAM.

\subsection{Prior Localization Test}
\subsubsection{EuRoC MAV Dataset}

EuRoC MAV Dataset \cite{euroc} is an indoor dataset containing 20Hz stereo camera images, 200Hz IMU, ground truth 6D pose, and ground truth laser scan data (prior LiDAR map). Hence, this dataset is suitable for the prior localization test. The prior LiDAR map is downsampled to a resolution of 5 cm. We assume that we have an accurate initial pose of MAV, and we simultaneously perform VIO localization and prior map-assisted localization.

Since our work is based on the state-of-the-art SLAM frame VINS-Fusion, we compare our work against the VINS-Fusion. Considering the evaluation metrics, we leverage Evo \cite{evo} to calculate the root mean squared error (RMSE) and standard deviation (STD) of the absolute trajectory error (ATE). To avoid randomness during feature tracking and keyframe selection, we perform each method five times on one trajectory and calculate the average RMSE and STD of ATE. Depending on the trajectory, our configuration slightly differs, with the related code available in our Github repository. The corresponding results are reported in Table \ref{euroc_ape}.

\AtBeginEnvironment{tabular}{\scriptsize}
\begin{table}[h]
        \setstretch{1.0}
        \caption{Average RMSE (m) and STD (m) of ATE on  EuRoC for PPCA-VINS and VINS-Fusion}
        \label{euroc_ape}
        \begin{center}
        \begin{tabular}{lcccc}
        \toprule
        \multicolumn{1}{c}{\multirow{2}{*}{Trajectory}} &
        \multicolumn{2}{c}{\textbf{PPCA-VINS}} & \multicolumn{2}{c}{\textbf{VINS-Fusion}} \\ \cmidrule{2-5} 
        \multicolumn{1}{c}{} & RMSE & STD & RMSE & STD  \\ \midrule[1px]
        V1\_01\_easy      & 0.1464 & 0.0197 & 0.1470 & 0.0231 \\ [2px]
        V1\_02\_medium    & 0.0904 & 0.0232 & 0.1170 & 0.0448 \\ [2px]
        V1\_01\_difficult & 0.1372 & 0.0393 & 0.1859 & 0.0576 \\ [2px]
        V2\_01\_easy      & 0.1785 & 0.1001 & 0.1790 & 0.1013 \\ [2px]
        V2\_02\_medium    & 0.1603 & 0.0678 & 0.1690 & 0.0760 \\ [2px]
        V2\_03\_difficult & 0.2387 & 0.1080 & 0.2445 & 0.1095 \\
        \bottomrule[1px]

        \end{tabular}
        \end{center}
\end{table}

Our PPCA-VINS performs better than VINS-Fusion on every trajectory ranging from easy to difficult. Hence, our method is more appealing than the standard VINS-Fusion affording a higher localization accuracy and a more stable trajectory error. However, our method's performance on trajectory V2\_03\_difficult is inferior to V1\_03\_difficult because the former trajectory is more dynamic and recorded in darker light conditions, limiting our point cloud generation scheme and creating more noise in our prior locate result.

\subsubsection{Change Detection Dataset}

We also compare our work to the standard VINS-Fusion on our dataset. Unlike the EuRoC dataset, our change detection dataset is collected in outdoor scenes, and thus we use the original scene's point cloud as the prior point cloud in PPCA-VINS. This means that we use an outdated point cloud as prior information.

\begin{table}[h]

        \caption{Small and Big Noise Parameters in Experiments}
        \label{noise_param}
        \begin{center}
        \begin{tabular}{ccc}
        \toprule
        noise & small & big \\
        \midrule
        $\sigma_g(rad\cdot s^{-1/2})$ & 4.0e-4 & 4.0e-3 \\
        $\sigma_a(m\cdot s^{-3/2})$ & 3.0e-3 & 3.0e-2 \\
        $\sigma_{bg}(rad\cdot s^{-3/2})$ & 4.0e-5 & 4.0e-4 \\
        $\sigma_{ba}(m\cdot s^{-5/2})$ & 3.0e-4 & 3.0e-3 \\
        $\sigma_{img}$ & 0.02 & 0.04 \\
        \bottomrule
        \end{tabular}
        \end{center}
\end{table}

\AtBeginEnvironment{tabular}{\scriptsize}
\begin{table}[h]

        \caption{Average RMSE (m), Standard Deviation (m), and Maximum (m) of ATE on our Dataset for PPCA-VINS and VINS-Fusion}
        \label{ch_ape}
        \begin{center}
        {\scriptsize
        \begin{tabular}{lcccccc}
                
        \toprule
        \multicolumn{1}{c}{\multirow{2}{*}{Trajectory/noise}} &
        \multicolumn{3}{c}{\textbf{PPCA-VINS}} & \multicolumn{3}{c}{\textbf{VINS-Fusion}} \\ \cmidrule{2-7} 
        \multicolumn{1}{c}{}  & RMSE   & STD    & MAX    & RMSE   & STD    & MAX \\ \midrule[1px]
            S1\_01/small      & 1.0969 & 0.5316 & 2.0758 & 1.7680 & 0.5029 & 2.8961\\ [2px]
            S1\_01/big        & 1.8654 & 0.8419 & 4.5178 & 2.7155 & 1.0426 & 5.2554\\ [2px]
            S1\_mirror/small  & 1.1999 & 0.6040 & 2.9738 & 2.1328 & 0.8652 & 4.3606\\ [2px]
            S1\_mirror/big    & 2.3591 & 1.1215 & 4.7216 & 3.5658 & 1.3468 & 6.2925\\ [2px]
            S1\_02/small      & 1.9786 & 1.0629 & 4.0199 & 3.0335 & 1.4665 & 5.8233\\ [2px]
            S1\_02/big        & 2.6830 & 1.3906 & 6.2569 & 4.2080 & 1.9974 & 8.6237\\ [2px]
            S2\_01/small      & 0.8584 & 0.4018 & 1.8421 & 1.0998 & 0.5024 & 2.5601\\ [2px]
            S2\_01/big        & 1.0623 & 0.5183 & 4.1442 & 2.3007 & 1.2102 & 5.4237\\ [2px]
            S2\_02/small      & 2.0957 & 1.0231 & 4.6308 & 2.8582 & 1.3670 & 5.3185\\ [2px]
            S2\_02/big        & 2.6681 & 1.4537 & 6.7778 & 3.3960 & 1.6929 & 7.4251\\ 
        \bottomrule[1px]

        \end{tabular}
        }
        \end{center}
\end{table}

{As our original dataset is noiseless, we add small noise and big noise to images (gaussian noise) and IMU readings (gaussian noise and random walk noise), parameters are shown in Table.\ref{noise_param}. Each trajectory is tested five times with both small and big noise, and the metrics are averaged over five runs.} The corresponding results are presented in Table \ref{ch_ape}, highlighting that even with an outdated prior point cloud, PPCA-VINS performs much better on each trajectory than the standard VINS-Fusion. {Moreover, the results reveal that our method runs successfully in both small noise and big noise conditions, and shows greater robustness than original VINS-Fusion. And for trajectory S1\_mirror, our PPCA-VINS's localization accuracy doesn't decrease much, but VINS-Fusion's accuracy has a big fall. Notice that for trajectory "S1\_dark", VINS-Fusion failed to run on it, so we will not anaylze this trajectory in following sections.}


\subsection{Change Detection Test}
In this section, we employ our suggested change detection dataset and set $th_{ch}$ to {3.2 m}, $\rho_o$ to {0.8 m}, and all point clouds are downsampled to 0.8m. 

\subsubsection{Free Space Ray Casting Test}
As mentioned in \ref{change-detection}, for pixels in filtered depth images that have an empty depth value, we cast a ray from the camera to depth $th_f$. This section evaluates various  $th_f$ depths to determine the optimal value.

\begin{figure}[h]
        \centering
        \includegraphics[width=0.5\textwidth]{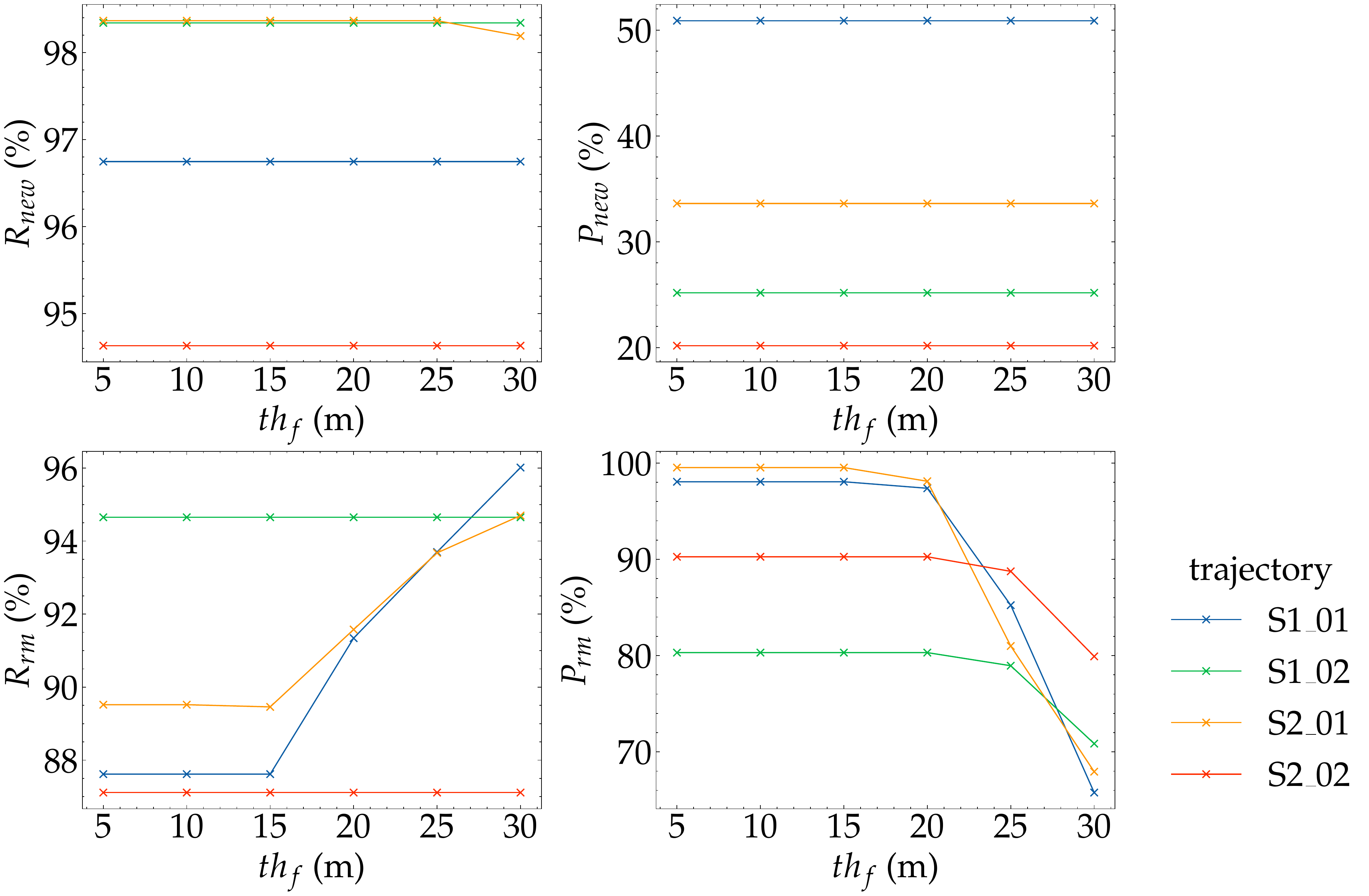}
        \caption{Recall and precision results of different classes when free space ray casting depth $th_f$ has various values.}
        \label{fig:metrics}
\end{figure}

We set $th_f$ to 5, 10, 15, 20, 25, and 30 m and calculate the recall and precision metrics in \ref{metrics} to determine the best  $th_f$ value. This experiment utilizes the 6D pose and the corresponding filtered depth image from PPCA-VINS with small sensor noise. Each metric is averaged over five runs. Fig.\ref{fig:metrics} illustrates the results revealing that $th_f$ has a great influence on $P_{rm}$. 

As presented in the figures above, setting $th_f$ to 15 or 20 meters attains the best performance. Thus, in the subsequent section we set $th_f$ to 20 meters.

\subsubsection{Comparison of Metrics using Different Pose Estimations}

This section calculates the metrics mentioned above with the 6D poses calculated from our method and the standard VINS-Fusion. The metrics are calculated for every trajectory and are averaged over five runs.

\begin{figure}[h]

        \centering
        \includegraphics[width=0.47\textwidth]{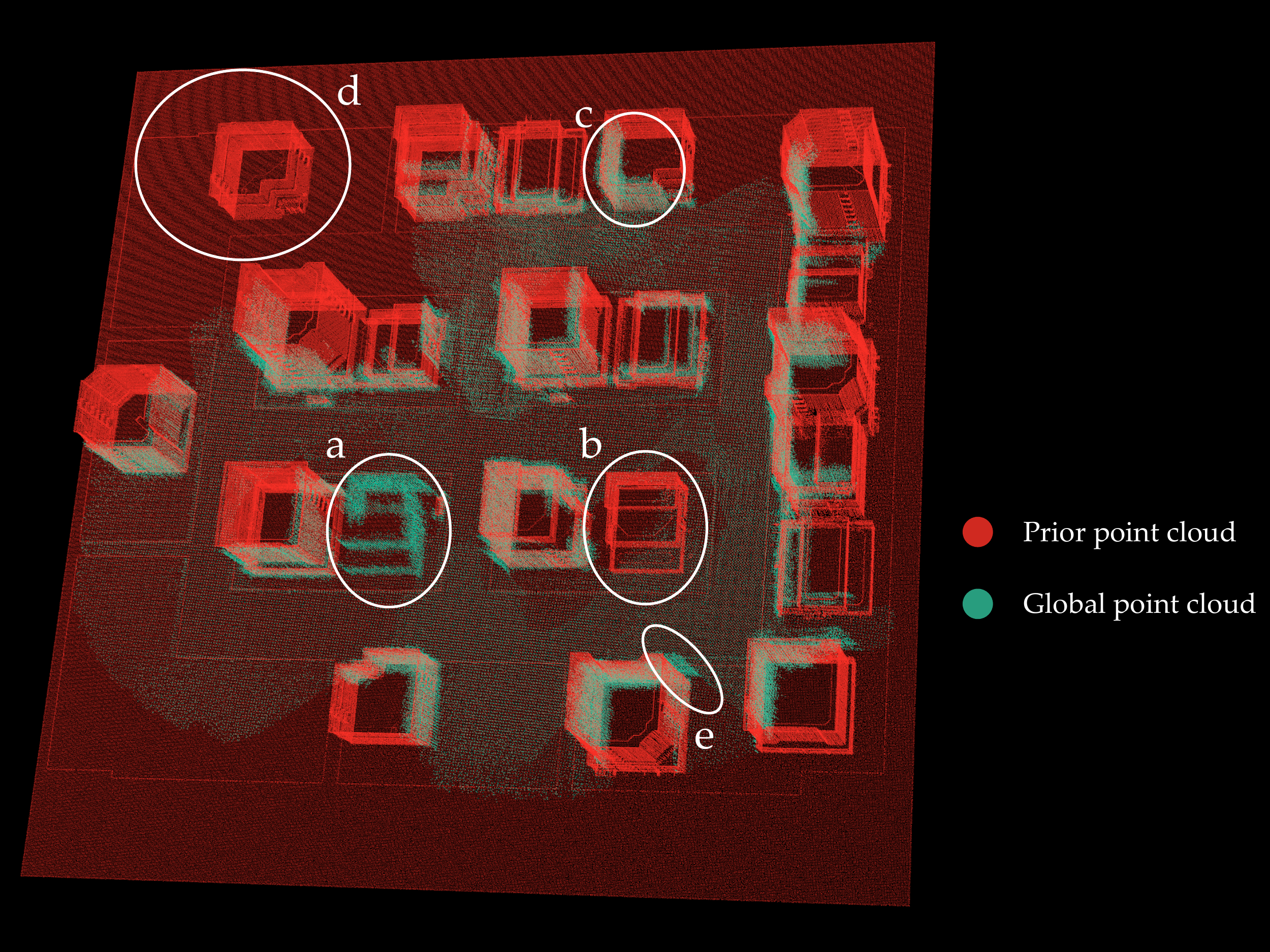}
        \caption{An illustration of different conditions when compare global point cloud to prior point cloud. a) new building; b) removed building; c) place where global point cloud and prior point cloud are aligned well; d) place where our MAV hasn't observed; e) place where global point cloud and prior point cloud are poorly aligned (Zoom in to have a clear view)}
        \label{fig:pc}
\end{figure}

In Fig.\ref{fig:pc}, we illustrate five different conditions when compare global point cloud to prior point cloud, thus making it easier to understand how these metrics are calculated.

\begin{figure}[h]

        \centering
        \includegraphics[width=0.5\textwidth]{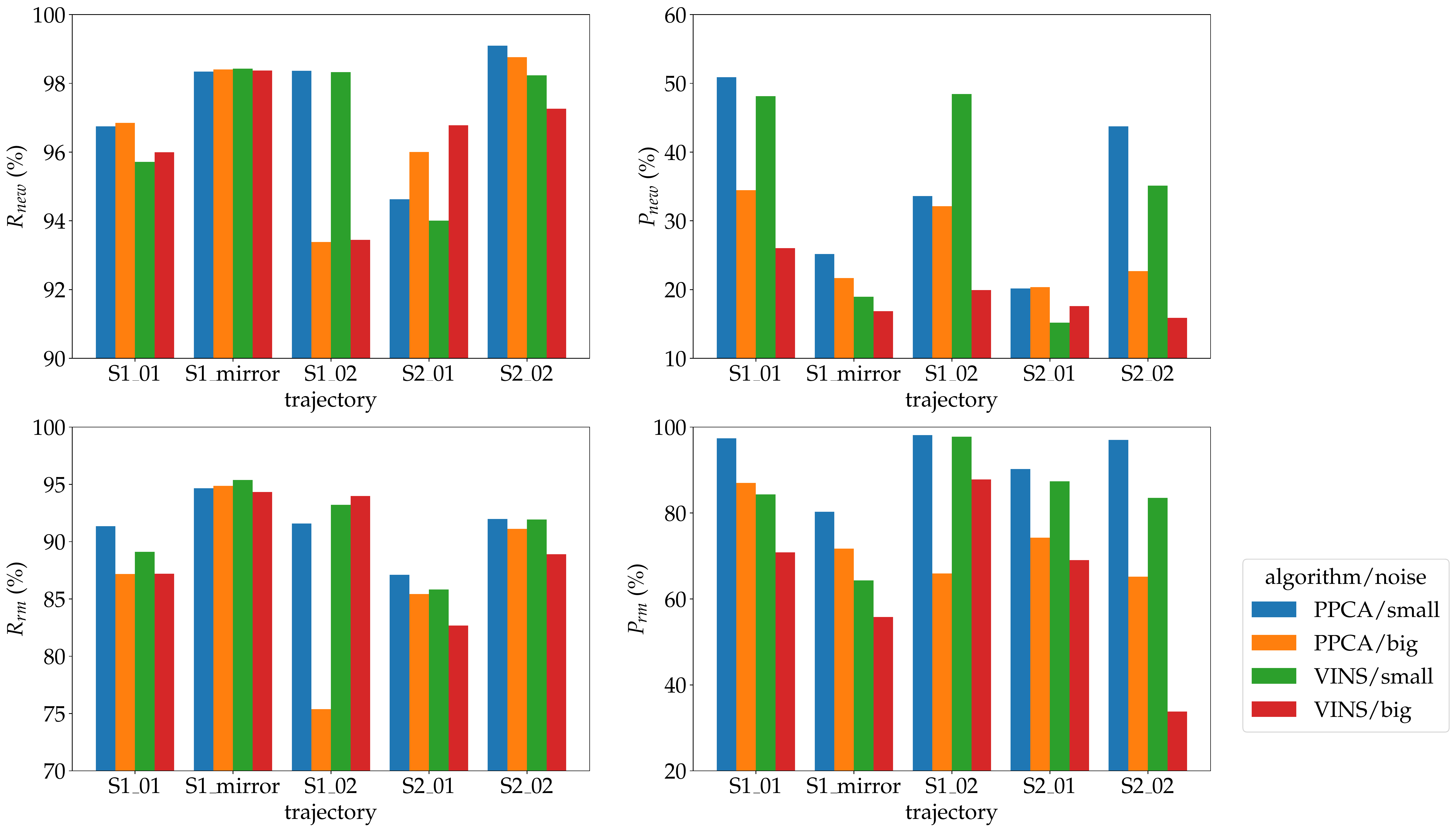}
        \caption{Comparison of the change detection results using 6D poses from PPCA-VINS and VINS-Fusion. }
        \label{fig:compare}
\end{figure}

Fig.\ref{fig:compare} highlights that PPCA-VINS affords a better performance on nearly every trajectory and metric, especially on $\mathcal{P}_{new}$ and $\mathcal{P}_{rm}$, indicating that our dataset and metrics are successful on point cloud change detection tasks. Besides, our visual-based baseline pipeline is verified effective on point cloud change detection tasks, affording a high recall on both new and removed points. Besides, we also found that when big noise is introduced to our trajectories, PPCA-VINS performs more robust than VINS-Fusion. Also, in scene S1\_mirror, $\mathcal{P}_{new}$ and $\mathcal{P}_{rm}$ is lower than that of scene without mirrors. This is because images with mirrors will produce wrongly estimated depth, thus making global point cloud has a poor quality.

Moreover, we find the $\mathcal{P}_{new}$ is significantly lower than $\mathcal{P}_{rm}$, for $\mathcal{P}_{new}$ is heavily influenced by the pose estimation accuracy. Indeed, when the estimated error increases, the re-projected point cloud from the camera to the real world drifts more, leading to a wrongly estimated observed area and imposing more re-projected points to be falsely classified as new points. $\mathcal{P}_{rm}$ is related to the prior point cloud and observed area. If the pose error is less than $th_{ch}$, the prior points in the wrongly observed area will also be within the threshold $th_{ch}$ of our global point clouds. Besides, in our scene, free space is much more than space occupied by buildings. So a minor drift very likely leads vehicles from observing a free area to observing another free space, which has a minor impact on the final metrics.

\subsection{Processing Time}
Our baseline framework is processing efficient, as on our PC, PPCA-VINS completes every trajectory of our dataset in 240 seconds (each scene in our dataset has a size of 250m $\times$ 250m $\times$ 40m). For the change detection pipeline, building point cloud having new and removed points requires less than five minutes per trajectory. Compared to Lidar-based and SfM-based methods, the proposed method has the advantages of relying on low-cost hardware and affords a low computational complexity. 

\section{CONCLUSIONS AND FUTURE WORK}
This paper proposes a novel visual-based point change detection pipeline and a simple simulated point change detection dataset and corresponding metrics. Compared to not employing a prior map, our PPCA-VINS achieves a higher pose estimation accuracy and better point cloud change detection results. Our visual-based point cloud change detection pipeline achieves an appealing result on our dataset, and the point cloud change detection metrics represent the corresponding performance efficiently.

We found that change detection results are significantly influenced by localization accuracy. Future work will improve localization accuracy using other kind of map (e.g. feature map). Moreover, we will conduct some real-world indoor tests.

\bibliographystyle{IEEEtran}
\bibliography{IEEEabrv.bib,ref.bib}
\end{document}